\documentclass[10pt, conference, letterpaper]{IEEEtran}
\IEEEoverridecommandlockouts
\usepackage{cite}
\usepackage{amsmath,amssymb,amsfonts}
\usepackage{graphicx}
\usepackage{textcomp}
\usepackage{xcolor}
\usepackage{comment}
\graphicspath{{figures/}}
\usepackage{mathtools}
\usepackage{float}
\usepackage{algorithm}
\usepackage{algorithmicx}
\usepackage[noend]{algpseudocode}
\usepackage{amsmath}

\DeclareMathOperator*{\argmin}{arg\,min}

\def\BibTeX{{\rm B\kern-.05em{\sc i\kern-.025em b}\kern-.08em
    T\kern-.1667em\lower.7ex\hbox{E}\kern-.125emX}}

\usepackage{tikz}
\newcommand\copyrighttext{%
  \footnotesize \textcopyright 2025 IEEE. Personal use of this material is permitted. 
  Permission from IEEE must be obtained for all other uses, in any current or future 
  media, including reprinting/republishing this material for advertising or 
  promotional purposes, creating new collective works, for resale or redistribution 
  to servers or lists, or reuse of any copyrighted component of this work in other works. Published in: 2024 IEEE 25th International Symposium on a World of Wireless, Mobile and Multimedia Networks (WoWMoM).  DOI: 10.1109/WoWMoM60985.2024.00041}
\newcommand\copyrightnotice{%
\begin{tikzpicture}[remember picture,overlay]
\node[anchor=south,yshift=10pt] at (current page.south) 
{\fbox{\parbox{\dimexpr\textwidth-\fboxsep-\fboxrule\relax}{\copyrighttext}}};
\end{tikzpicture}%
}

\begin{document}

\title{NaviSplit: Dynamic Multi-Branch Split DNNs for Efficient Distributed Autonomous Navigation
}

\author{
\IEEEauthorblockN{1\textsuperscript{st} Timothy K Johnsen}
\IEEEauthorblockA{
\textit{University of California Irvine, USA}\\
tjohnsen@uci.edu}
\and
\IEEEauthorblockN{2\textsuperscript{nd} Ian Harshbarger}
\IEEEauthorblockA{
\textit{University of California Irvine, USA}\\
iharshba@uci.edu}
\and
\IEEEauthorblockN{3\textsuperscript{rd} Zixia Xia}
\IEEEauthorblockA{
\textit{University of California Irvine, USA}\\
zixiax3@uci.edu}
\and
\IEEEauthorblockN{4\textsuperscript{th} Marco Levorato}
\IEEEauthorblockA{
\textit{University of California Irvine, USA}\\
levorato@uci.edu}
}

\maketitle
\copyrightnotice

\begin{abstract}
Lightweight autonomous unmanned aerial vehicles (UAV) are emerging as a central component of a broad range of applications. However, autonomous navigation necessitates the implementation of perception algorithms, often deep neural networks (DNN), that process the input of sensor observations, such as that from cameras and LiDARs, for control logic. The complexity of such algorithms clashes with the severe constraints of these devices in terms of computing power, energy, memory, and execution time. In this paper, we propose NaviSplit, the first instance of a lightweight navigation framework embedding a distributed and dynamic multi-branched neural model. At its core is a DNN split at a compression point, resulting in two model parts: (1) the head model, that is executed at the vehicle, which partially processes and compacts perception from sensors; and (2) the tail model, that is executed at an interconnected compute-capable device, which processes the remainder of the compacted perception and infers navigation commands. Different from prior work, the NaviSplit framework includes a neural gate that dynamically selects a specific head model to minimize channel usage while efficiently supporting the navigation network. In our implementation, the perception model extracts a 2D depth map from a monocular RGB image captured by the drone using the robust simulator Microsoft AirSim. Our results demonstrate that the NaviSplit depth model achieves an extraction accuracy of 72-81\% while transmitting an extremely small amount of data (1.2-18 KB) to the edge server. When using the neural gate, as utilized by NaviSplit, we obtain a slightly higher navigation accuracy as compared to a larger static network by 0.3\% while significantly reducing the data rate by 95\%. To the best of our knowledge, this is the first exemplar of dynamic multi-branched model based on split DNNs for autonomous navigation.
\end{abstract}


\section{Introduction}
\label{sec:introduction}

Autonomous navigation increasingly relies on the execution of computationally expensive deep neural networks (DNN) that integrate perception tasks (\textit{e.g.,} object detection, segmentation, or depth estimation) with decision making tasks. In some settings (\textit{e.g.,} lightweight airborne drones), the limited onboard hardware can complicate execution of such DNNs. Such applications impose severe execution deadlines needed to improve reaction time to the surrounding environment. Thus, it is beneficial to develop modern solutions that allow execution of DNN models on lightweight autonomous vehicles while minimizing execution time, memory, and energy expenditure.

Typical solutions adopt two approaches. \textbf{Model reduction} simplifies DNN models to be executed onboard, such as: quantization \cite{pappalardo2022qonnx} \cite{banner2018scalable} \cite{courbariaux2015training}, knowledge distillation \cite{hinton2015distilling} \cite{ahn2019variational}, \cite{mishra2017apprentice}, and direct design \cite{gholami2018squeezenext}. The resulting models incur a degradation in task performance, and their continuous execution requires considerable energy expense. \textbf{Edge computing} \cite{EgdeComputing-SmartCities,VideoEdge} remotely executes the full DNN at a compute-capable device -- the edge server. This transmits input data (\textit{e.g.,} images) over volatile and capacity-constrained wireless links, creating problems in efficient channel usage, delay, delay variance, and security.

We focus on edge computing for lightweight autonomous unmanned aerial vehicles (UAV) (\textit{e.g.,} micro-drones), where hardware limitations affect both computing and sensing capacities. We consider the task of navigating a UAV to a target position through an unknown environment. The UAV is equipped with an efficient monocular RGB camera, which provides limited information toward navigation. Path planning and collision avoidance requires information on the 3D structure of the surrounding space. Thus, the autonomy pipeline is composed of: (1) a DNN that extracts depths from an input image; and (2) a DNN that inputs extracted depths and state information to output navigation motions.

A straightforward application of edge computing would require the vehicle to transmit the (compressed) image to the edge server, which then executes both the depth and navigation DNNs. Such an exchange imposes considerable channel usage, while exposing the control loop to latency variations due to erratic capacity patterns typical in airborne systems \cite{seremas}.

Herein, we propose an innovative architecture -- \emph{NaviSplit} -- that uses split DNNs and supervised compression \cite{matsubara2022split} to build a dynamic and efficient, distributed navigation framework. First, we create a ``bottleneck'' \cite{matsubara2022bottlefit} within the depth DNN, where a lower-dimension tensor representation is trained to support the supervised task. This creates more robust representations than that of typical autoencoders, which are trained to simply reconstruct the input image. Second, the portion of the model up to the bottleneck (head model) is executed  onboard the vehicle, and the rest of the model (tail model) is executed at the edge server. We note that our design is the first example of supervised compression for depth estimation. Third, we train several split DNN models that range in computational complexities, compression rate, and depth accuracy -- then encapsulate these models in a gated dynamic network framework. Finally, we train an auxiliary model to select the split DNN that minimizes channel usage while supporting navigation. The resulting architecture tunes the computing requirements, channel usage, and depth estimation accuracy, to the changing navigational needs of the vehicle.

%Then, we encapsulate the split DNN model in a gated dynamic network framework. Specifically, we create an auxiliary model, that takes as input recent depth maps generated by the split network and selects a specific head model among a set designed to have offer a range of computational complexities and compression rate (and depth map accuracy). The auxiliary model is trained to support the navigation control while minimizing channel usage. The overall architecture, then, tunes the computing effort, channel usage and depth estimation accuracy to the specific current navigational needs of the vehicle.

The main contributions of this paper are as follows.

\vspace{0.5mm}
\noindent
$\bullet$ We present \emph{NaviSplit}, an adaptable multi-branched neural architecture with supervised compression. Our core innovations are: (a)  a novel distributed neural architecture, and (b) a novel multi-stage training process that results in an auxiliary model that dynamically selects optimal compression factors. 
%In short, we first train several task models with various splitting schemes, and then use reinforcement learning to train an auxiliary model that selects from those head models to accomplish navigation objectives while minimizing data transfer.

\noindent
$\bullet$ We implement \emph{NaviSplit} on the robust simulator Microsoft AirSim \cite{shah2018airsim}, where the task is to navigate a drone to a target location while minimizing path length and avoiding collisions.

%The task model utilizes large CNNs that transform input RGB images into depth maps and further into drone motion actions used for navigation.

\noindent
$\bullet$ We release our simulation tool, that interfaces with Microsoft AirSim, open-source to the public.

Our results demonstrate that the \emph{NaviSplit} depth DNN achieves an extraction accuracy of 72-81\% while transmitting an extremely small amount of data (between 1.2 and 18 KB) to the edge server. Compared to a static state-of-the-art (SoA) model, \emph{NaviSplit} obtains a slightly higher mean navigation accuracy (82.5\% versus 82.2\%) with a mean reduction in 95\% transmitted data (43 kilobyte/meter versus 2.1 KB/m). 

%To the best of our knowledge, this is the first exemplar of dynamic multi-branched model based on split DNNs for autonomous navigation.

\section{Related Work}
\label{sec:related}

In edge computing, data is collected by a mobile device and transferred to a compute-capable edge server over a wireless channel. The server processes the data and in some settings relays them back to the mobile device \cite{EgdeComputing-SmartCities}. Processing typically consists of executing DNNs to extract features, semantics, and control. In computer vision applications such as \cite{ECDNNImageRecognition}, the mobile device can apply JPEG or MPEG compression to compress input images and reduce the amount of data. 

Split computing (SC)~\cite{matsubara2022split}, also known as split DNN and model partitioning, is a recent class of approaches in mobile computing. DNN architectures are partitioned into two sections -- head and tail -- that are executed by the mobile device and edge server, respectively. The objective is to balance computing load, energy consumption, and channel usage. While early approaches \cite{NeuroSurgeon} simply ``split'' existing DNN models, recent methods involve injecting a "bottleneck" into a trained DNN task model. This alters and trains the DNN's layers to learn a compact set of task-specific features that preserves the task accuracy \cite{matsubara2022bottlefit}.

A recent approach allowed for dynamic quantization of intermediate activation values with fixed activation sizes at the split point \cite{assine2023slimmable}. However, the literature is sparse in approaches to dynamically control the size of the activation values as transmitted at the split point. Herein, we design a gated neural architecture that can dynamically select the compression model given perceived context. We remark how, to the best of our knowledge, this is the first exemplar of such construction, as well as the first application of split computing to both a depth estimation and navigation problem.

\section{Methods}
\label{sec:methods}

\begin{figure}[htbp]
\centerline{\includegraphics[width=0.4\textwidth]{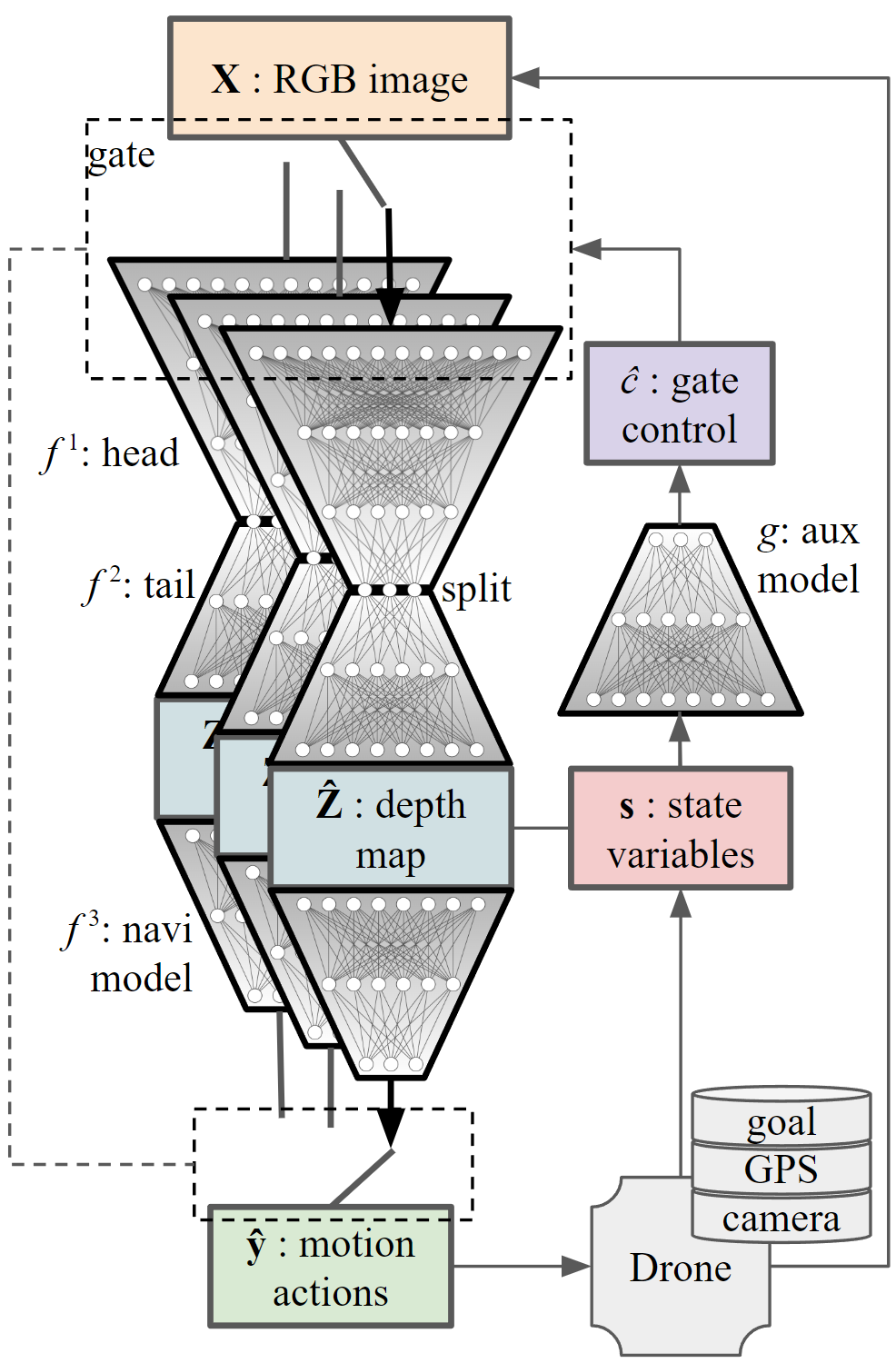}}
\caption{The framework we propose uses a teacher model that maps a monocular RGB image to a depth map. Several different split points with encoder/decoder are injected into the teacher model to make multiple student model branches capable of split computing -- of which an auxiliary model selects from given perceived context. Extracted depth maps are input to the navigation model that outputs motion actions used during drone navigation.}
\label{fig:sys_model}
\end{figure}

Figure~\ref{fig:sys_model} illustrates the system model for \emph{NaviSplit}. We consider a system composed of: a lightweight UAV equipped with a monocular forward-facing RGB camera and small computing chip such as a Jetson Nano or Raspeberry Pi, and an edge server equipped with significantly more computing resources. The objective of the drone is to navigate to a target position in an unknown environment while minimizing path length and avoiding collisions. To this aim, the images captured by the onboard camera are first transformed into a depth map by a depth DNN. Second, the extracted depths are combined with state variables (\textit{e.g.,} current and target positions) and then fed to a navigation DNN to produce motion commands. Given the limited resources available to the drone, we assume that it is either impossible or inefficient to execute the control pipeline at the drone, for instance due to memory constraints, insufficient energy availability, or excessive latency. Thus, we split the depth DNN into a head model that is computed onboard and a tail model that is computed at the edge server. The most novel component of the presented system is the auxiliary model that controls the gate used to select from the multi-branch split DNN framework.

\vspace{1mm}
\noindent
{\bf Edge computing:} a viable option is for the drone to compress the images captured by the camera (e.g., using JPEG), that are then used by the edge server to execute the pipeline transforming images to motion commands.
However, especially in systems with extreme resource constraints -- e.g., a nano drone connected to a mobile base station, the wireless link connecting the drone to the edge server may have a severely constrained capacity, where the achievable data rate has an erratic pattern due to the motion characteristics of the drone.

\subsection{NaviSplit Approach}

We seek a methodology to reduce the amount of data transferred over the channel while preserving navigation performance. To this aim, \emph{NaviSplit} develops a new generation of neural models combining split computing and supervised compression with a gated multi-branched model. The gate is driven by a specialized auxiliary module to select encoder/decoder pairs built using a supervised compression approach. The rationale is to select a compression strategy matching the needs of the controller, that is, capable of producing representations suitable to determine control given the operating context. There are five modules composing \emph{NaviSplit}.

\vspace{1mm}
\noindent
$\bullet$ {\bf Sensing (Camera):} we collect sensor data (RGB images) that is sufficient to fulfil mission objectives as accomplished by the downstream task (navigation) model. We use the notation \textbf{X} to refer to an observation acquired from onboard sensor(s).

\vspace{1mm}
\noindent
$\bullet$ {\bf Depth Maps:} in our implementation, the acquired sensor data is transformed into an intermediate data structure taking the form of a 2D depth map, $\hat{\mathbf{Z}}$, which is a representation of the relative distances between the drone and nearby objects within the field of view of onboard sensors.

\vspace{1mm}
\noindent
$\bullet$ {\bf Task (Navigation):} sensor data and state variables, $\mathbf{s}$, are transformed into task output, $\hat{\mathbf{y}}$. In our implementation, $\hat{\mathbf{y}}$ contains motion actions for the drone to navigate between its current and target location, sensor data is transformed into $\hat{\mathbf{Z}}$, and $\mathbf{s}$ contains the current and target GPS positions.

\vspace{1mm}
\noindent
$\bullet$ {\bf Split Computing:} several sensing-depth-navigation student models are created which each use a different split computing design. This results in a spectrum of models to select from, with various encoded data sizes that are indexed by a gate control value, $c$. We remark how the drone only needs to store and execute (one per image) the head portion of the model (the encoder), which is built to be of minimal complexity.

\vspace{1mm}
\noindent
$\bullet$ {\bf Adaptation:} An auxiliary model, $g_\phi$, is developed to intelligently select an encoder/decoder pair, in response to perceived context at the current time step. To the best of our knowledge, this is the first application of dynamic, adaptive split computing and is the core contribution of this paper.
\vspace{1mm}

We aim to solve the following optimization problem:
\begin{equation}
\label{eq:opt1}
\begin{aligned}
\argmin_{\phi} \; \left\langle \hat{c} \right\rangle \\
\textrm{s.t.} \; \left\langle \eta\left(g_\phi, \{\mathbf{s}^{(t=0)}\}\right) \right\rangle \geq \beta * \left\langle \eta\left(c_{max}, \{\mathbf{s}^{(t=0)}\}\right) \right\rangle
\end{aligned}
\end{equation}
Quantity $\{\mathbf{s}^{(t=0)}\}$ denotes a set of training objectives defined by their initial state variables (starting and target positions). Each value of $\mathbf{s}^{(t=0)}$ will result in a path taken by the drone as decided by both the multi-branch navigation DNN, $f$, and auxiliary model, $g_\phi$, such that one path will be executed using:

\begin{equation}
\label{eq:f}
\begin{aligned}
\hat{\mathbf{y}}^{(t)} = f\left(\mathbf{X}^{(t)}, \mathbf{s}^{(t)}, \hat{c}^{(t)}\right) \\ 
\hat{c}^{(t)} = g_\phi\left(\hat{\mathbf{Z}}^{(t-1)}, \mathbf{s}^{(t-1)}\right) \\
\hat{c}^{(t=0)} = c_{max} \\
\hat{\mathbf{Z}}^{(t)} = f_2\left(f_1\left(\mathbf{X}^{(t)}, \hat{c}^{(t)}\right), \hat{c}^{(t)}\right) \\ 
\end{aligned}
\end{equation}

where time steps will be computed until a termination criteria is met. Thus, $\left\langle \hat{c} \right\rangle$ denotes the expected value of the gate control -- given that a lower indexed value of $c$ correlates to a smaller data rate used in SC, and $c_{max}$ is the maximum gate value corresponding to the $f_\theta \in f$ with the largest data rate. Quantity $\eta$ is the task accuracy which is a function of the initial state variables and the auxiliary mechanism, where the auxiliary mechanism that always uses $c_{max}$ corresponds to the SoA static case, and $\beta$ is a scalar parameter set by the user. A smaller data rate should result in lower accuracy, and thus a typical parameter range is $0 < \beta \leq 1$. Specific to our implementation, we consider $\eta$ to be the navigation accuracy (\textit{i.e.}, percent of successful paths). Where success is measured by reaching the target position, while avoiding collisions, and within a maximum number of time steps.

The subsequent sections detail each step of a multi-stage procedure we use to train the several model parts of \emph{NaviSlim}. We implement our approach on the robust drone simulator Microsoft AirSim \cite{shah2018airsim}, which is integrated with our open-source Python interface to train and evaluate models with\footnote{https://github.com/WreckItTim/rl\_drone}.

\subsection{Depth Model}
\label{sec:depth}

Depth maps provide rich information used to calculate precise movements. Given a monocular RGB camera, typical in small drone applications, these depths can not be directly calculated and instead must be extracted. We use a convolutional neural network (CNN) to transform $\mathbf{X} \to \hat{\mathbf{Z}}$, by minimizing the expected $L_1$-loss extraction error:

\begin{equation}
\label{eq:opt2}
\begin{aligned}
\argmin_{\theta} \; \left\langle L_1\left(\mathbf{Z}, f_{p,\theta}\left(\mathbf{X}\right) \right) \right\rangle 
\end{aligned}
\end{equation}
where we use $f_{p,\theta}$ to refer to the parent depth model -- which will later be split into student models. The ground truth depth maps, $\mathbf{Z}$, are obtained directly from AirSim by creating datasets of known mappings $\mathbf{X} \to \mathbf{Z}$, including 4500 training images and 500 testing images. The shape of an RGB image is [3, 144, 256]; and the shape of a depth map is [144, 256].

The depth model consists of ten feature extraction blocks followed by a depth prediction block. Every feature extraction block includes a convolution layer, a group normalization layer~\cite{Wu_2018_ECCV}, and a scaled exponential linear unit (SELU)~\cite{klambauer2017self} activation layer. The depth prediction block includes three convolution layers, one group normalization layer, two SELU activation layers, and a Sigmoid activation layer -- where a value of one corresponds to over 100 meters away and a value of zero is directly in front of the camera. We train each depth model using an Adam optimizer \cite{kingma2014adam} and learning rate decay.

\subsection{Split Computing}
\label{sec:split}

We evaluate two methods of injecting a compressed split point into $f_{p,\theta}$. The first is a baseline model that simply reduces the number of channels between two blocks around the split point. The second is a more advanced method similar to that presented in \cite{matsubara2022bottlefit}, which injects a bottleneck around the split point that changes the structure of multiple blocks.

For the baseline student models, we reduce the number of output channels in the second block of $f_{p,\theta}$ down from 128 to either 2, 4, 8, 16, or 32, which subsequently reduces the number of input channels in the third block. In the following discussion, $\mathbf{l}_t = \{1\}\cup\{4, ..., 11\}$, $\mathbf{l}'_t = \{3, ..., 11\}$, and $\mathbf{l}_ h= \{2, 3\}$ -- which denote sets of block indices.

For the bottleneck student models, we design two custom bottlenecks by creating encoder and decoder sections. We target the decoder's reconstruction to fit the fifth convolution block output from $f_{p,\theta}$. Within the bottleneck, we compress the height and width to a size of either 4x9 (which we refer to as Bottleneck\_V1) or 8x14 (Bottleneck\_V2). Further, we compress the channel values down from 64 to either 12, 24, or 48. The structure of the injected bottlenecks can be viewed as an entirely separate model replacing the first five convolution blocks of $f_{p,\theta}$. In the following discussion, $\mathbf{l}_t = \{6, ..., 11\}$, $\mathbf{l}'_t = \{5, ..., 11\}$, and $\mathbf{l}_ h= \{1, ..., 5\}$.

\textbf{Head Training.} 
The trainable parameters for blocks on indices $\mathbf{l}_t$ (corresponding to the tail) are directly copied over from the teacher model, where as those for $\mathbf{l}_h$ (corresponding to the head) are randomly initialized. The trainable parameters for blocks at indices $\mathbf{l}_t$ are frozen, and our encoder/decoder is trained as follows. We use knowledge distillation \cite{hinton2015distilling} so that the error gradient is calculated with respect to the output of blocks at indices $\mathbf{l}'_t$ in the student model against those in $f_{p,\theta}$. We use an $L_2$-loss function applied to each layer, an Adam optimizer, learning rate decay, and early stopping \cite{prechelt2002early}. The error gradient is propagated and used to update the trainable parameters from blocks at indices $\mathbf{l}_h$:

\begin{equation}
\label{eq:head}
\begin{aligned}
\nabla = \frac{\partial}{\partial \theta'} \Sigma \{ L_2\left(f_{p,\theta}^{(i)}\left(\mathbf{X}\right), f_{s,\theta'}^{(i)}\left(\mathbf{X}\right)\right) \; \forall \; i \in \mathbf{l}'_t \} \\
\end{aligned}
\end{equation}
where $f_{s,\theta'}$ is the student model, and $f_\cdot^{(i)}$ denotes the $i^{th}$ block of $f_\cdot$ (either teacher or student model) from the set $\mathbf{l}'_t$.

\textbf{Tail Training.} 
The trainable parameters in the injected bottleneck, blocks at indices $\mathbf{l}_h$, are then frozen and those in the blocks at indices $\mathbf{l}_t$ are unfrozen for fine tuning as follows. We calculate the error gradient using an $L_2$-loss on the sum of two terms: the difference between ground truth task output (otherwise called a ''hard" target) and the difference between the student and teacher output (otherwise called a ''soft" target). The error gradient is propagated and used to update the trainable parameters from blocks at indices $\mathbf{l}_t$:

\begin{equation}
\label{eq:head}
\begin{aligned}
\nabla = \frac{\partial}{\partial \theta'} [L_2\left(f_{p,\theta}\left(\mathbf{X}\right), f_{s,\theta'}\left(\mathbf{X}\right)\right) + L_2\left(\mathbf{Z}, f_{s,\theta'}\left(\mathbf{X}\right)\right)]\\
\end{aligned}
\end{equation}

To further reduce the size of encoded data (in kilobytes) before communicating with an edge server, we quantize the compressed encoding to an 8-bit unsigned integer tensor. This way, the memory transmitted is similar to that of JPEG compression of the input RGB image, where one of our 32-channel baseline student models would communicate a tensor of size 18.4 KB as compared to a JPEG compression with 95 quality that would communicate a mean size of 20.6 KB.

\subsection{Navigation Model}
\label{sec:navigation}

The navigation model is tasked to transform $\mathbf{\hat{Z}}$ to $\mathbf{y}$. We aim to keep the size of the navigation model minimal, thus preprocess $\mathbf{\hat{Z}}$ by applying a min pooling layer that reduces the depth map to a size of 8x6, which is then flattened into a vector. We append the relative difference between current and target positions, such that $\mathbf{s} = [\Delta x, \Delta y, \Delta z, \Delta \text{yaw}]$. Data, $\mathbf{\hat{Z}}$ and $\mathbf{s}$, from the four most recent time steps are concatenated in temporal order. This results in a feature vector of length 208, which is fed into a MultiLayer Perceptron (MLP) with 3 layers of 32 Rectified Linear Unit (ReLU) \cite{agarap2018deep} nodes each, followed by an output layer that is squashed with hyperbolic tangent. These output nodes control drone translation and rotation. The navigation neural network is then trained using a TD3 reinforcement learning algorithm \cite{fujimoto2018addressing}, utilizing the StableBaselines-3 (SB3) \cite{raffin2021stable} python library with Pytorch \cite{paszke2017automatic}.  This approach is similar to SoA static approaches \cite{b18, b6, b15, b17}.

We use the following reward function:

\begin{equation}
\label{eq:reward}
r(t) {=} \left\{
\begin{array}{ll}
      - \alpha_1 & collision \\
      \alpha_2 & goal \\
      \alpha_3 tanh(d) {-} \alpha_4  & intermediate \\
\end{array} 
\right. 
\end{equation}
where the $\alpha$ terms are positive constants set by the user. The first condition applies a large penalty for colliding with an object, and the second condition applies a large reward for reaching the target position. The third condition is used during intermediate time steps, and applies a reward for moving closer to the target position by using the relative distance, $d$, and applies a constant time penalty to encourage shorter paths. Further, we apply a curriculum learning schedule to incrementally increase the difficulty. One navigation model is trained for, and paired with, each student depth model.

\subsection{Auxiliary Model}
\label{sec:auxiliary}

The objective of the auxiliary model, $g_\phi$, is to select $f_\theta \in f$ by inferring the gate control value, $\hat{c}$. We use an MLP with 2 ReLU layers of 32 hidden nodes each. The output layer is a hyperbolic tangent node which outputs a normalized value of $\hat{c}$. As input, we append the four most recent values of $\hat{c}$ to the same feature vector used as input into the navigation model.

We train the auxiliary network $g_\phi$ using similar methods as the navigation network --  a TD3 reinforcement learning algorithm with a similar reward function as outlined in Equation~\ref{eq:reward} and curriculum learning schedule. Only, we add a penalty for high values of $\hat{c}$ to the intermediate condition, such as: $-\alpha_5\hat{c}$.

\section{Results}
\label{sec:results}

We evaluate the teacher depth model used to extract depths from a monocular RGB image. As expected, objects further away receive a higher extraction error. There is a linear increase in the root mean squared error (RMSE) from 9.4 to 21.2 meters, as the ground truth depth bins range from 10 to 90 meters. RMSE drops to 16.1 meters when extracting depths at the horizon (those clipped to 100 meters away).

\begin{figure}[htbp]
\centerline{\includegraphics[width=0.4\textwidth]{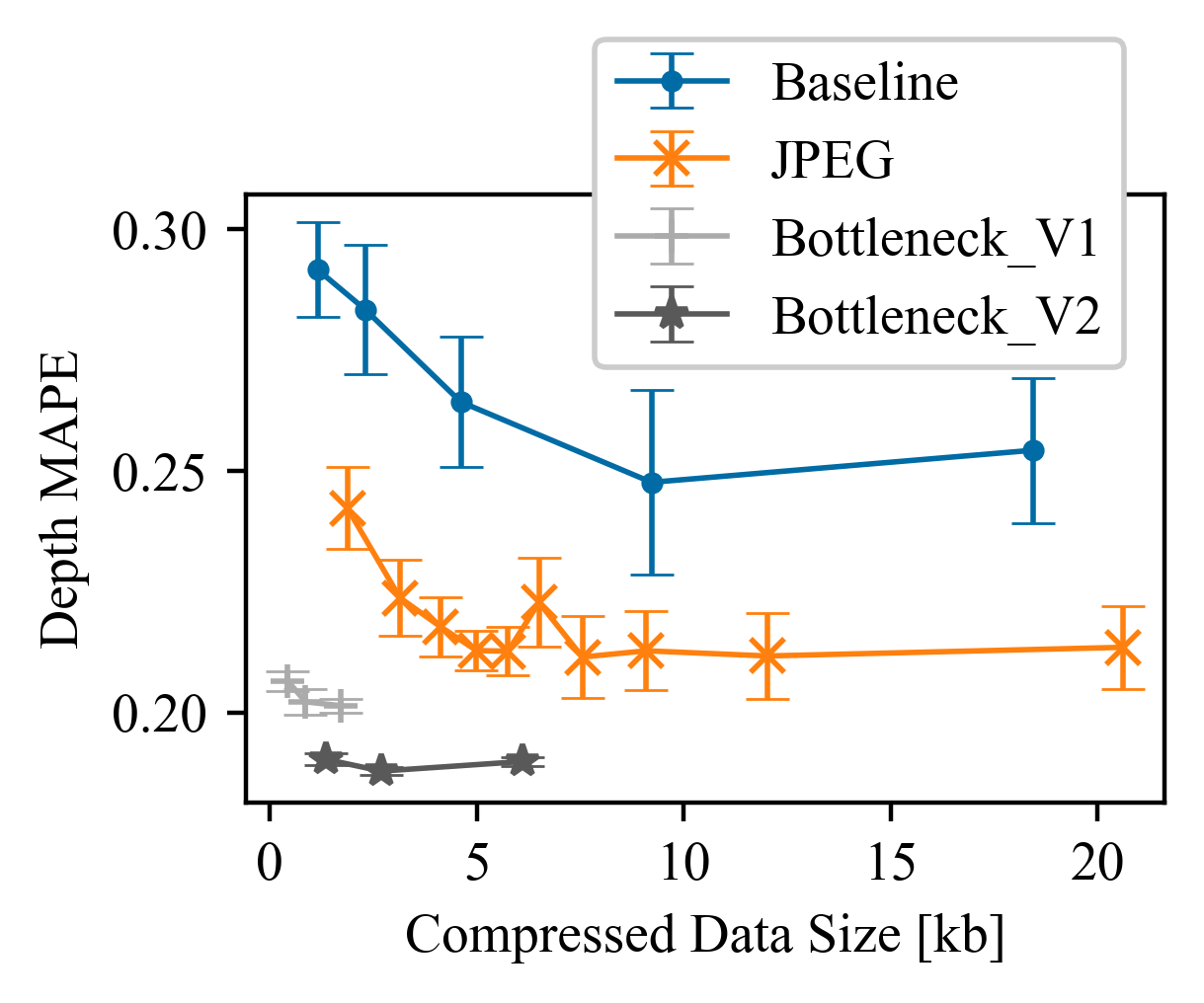}}
\caption{Comparing various compressed data sizes, corresponding to different models, versus resulting depth extraction error. The markers from left to right: for bottleneck models, range between a reduction in channels of [12, 24, 64]; for baseline models, range between a reduction in channels of [2, 4, 8, 16, 32]; and for JPEG models, range in quality of compression from 5 to 95.}
\label{fig:split_err2}
\end{figure}

Figure~\ref{fig:split_err2} compares the baseline student models, the bottleneck student models, and edge computing that completely offloads the image using JPEG compression. We see that the baseline models perform with equal memory consumption as both the bottleneck and JPEG models, however do not result in a lower error than JPEG. Alternatively, the bottleneck models perform with better error than that of the JPEG models - warranting the bottleneck methodology is more robust.

Using linear quantization slightly improves the depth extraction error, ranging in a decrease of mean absolute percent error (MAPE) of 0.8 to 4\% -- improving with increased compression. We assume this is due to some level of regularization due to the mapping of a 32-bit floating point number to an 8-bit unsigned integer, along with inherent clipping (eliminating possible outliers that may otherwise blow up towards infinity).

Using the student models to extract depth maps, we then train and evaluate each of the navigation models. For navigation accuracy, we use the percent of successful paths -- where a successful path is one that reached the target position, without any collisions, and within a maximum number of computational time steps. We evaluate each navigation model against a static set of initial and target positions. We find that the navigation accuracy has no relationship with the distance between initial and target positions -- as this is instead a function of the complexity of the path (number and size of objects in the way). However, the number of computational time steps required to execute a successful path has a linear relationship with initial distance -- thus is a good normalization value for benchmarks. The auxiliary model is trained to select from these sensing-depth-navigation student model branches. 

Table~\ref{tab:results} lists the following benchmarks for the teacher, student, and auxiliary models: navigation accuracy, the normalized number of time steps per meter of initial distance, and the normalized compressed data size (communicated during SC) per meter of initial distance. For the two normalized benchmarks, only the successful paths are considered because unsuccessful ones terminate either early from collision or late from the max time step requirement. Normalization is needed, because different student models result in a different set of evaluation paths that successfully reach the target position.

\begin{table}[htbp]
\caption{}
\label{sensors}
\begin{center}
\begin{tabular}{|c|c|c|c|}
\hline
\textbf{Model} & \textbf{Navigation [\%]} & \textbf{Path [ts/m]} & \textbf{Encoded [KB/m]} \\ 
\hline
Teacher (SoA) & 82.2 & 0.1446 & 42.64 \\
\hline
Student-1 & 81.6 & 0.1487 & \textbf{0.6856} \\
Student-2 & 78.6 & 0.1504 & 2.772 \\
Student-3 & 79.3 & \textbf{0.1436} & 10.58 \\
Auxiliary & \textbf{82.5} & 0.1489 & 2.12 \\
\hline
\end{tabular}
\label{tab:results}
\end{center}
\end{table}

Using the auxiliary model consistently results in a higher navigation accuracy as compared to using the student models independently, and in fact receives a higher navigation accuracy than when using just the SoA teacher model. Further, the auxiliary model obtains this navigation accuracy using a substantially smaller encoded feature space. The only model which was executed using a smaller encoded space received a lower navigation accuracy due to it being over-compressed. The benefit of \emph{Navisplit}, with an auxiliary model, is being able to select from several student models -- thus improving navigation accuracy while using a dynamic encoded data size.

\begin{figure}[htbp]
\centerline{\includegraphics{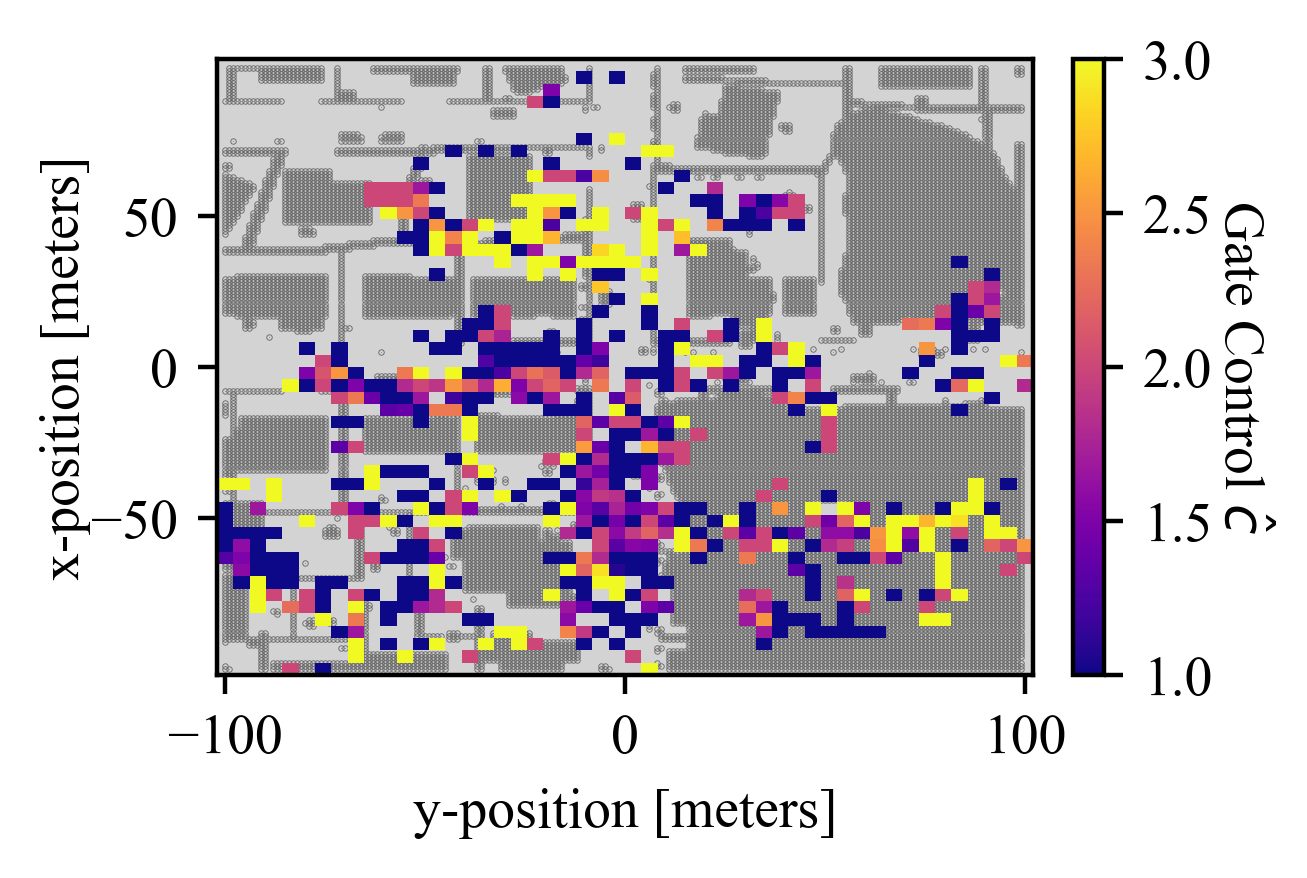}}
\caption{Mean value of the gate control, $\hat{c}$, for all successful paths while using an auxiliary model to adapt and select $\hat{c}$ at each time step. This a demonstration of \emph{NaviSplit}, which was trained and evaluated in Microsoft AirSim.}.
\label{fig:heats1}
\end{figure}

To evaluate the behavior of \emph{Navisplit}, we consider all successful evaluation paths flown while using the auxiliary model. The gate control index, $c$, for each student model increases in magnitude with increasing compressed data size. Figure~\ref{fig:heats1} shows the mean gate control value predicted by the auxiliary model, $\hat{c}$, at each position. The apparent learned behavior from the auxiliary model is to use the smallest student model as often as possible -- reducing the size of compressed data at the split point -- and only activating more expensive student models as needed. Thus, the auxiliary model effectively increases navigation accuracy while minimizing the compressed data size used throughout the path. We see a low value of $\hat{c}$ for open roads and a high value of $\hat{c}$ when navigating around homes and other objects - warranting that a larger encoded data size is needed for more difficult scenarios.

\section{Conclusions}
\label{sec:conclusions}

We have presented \emph{NaviSplit} -- an effective multi-branched neural network architecture for drone navigation, that dynamically splits computing between an equipped processing unit and edge server. \emph{NaviSplit} slightly improved navigation accuracy by 0.3\% over a larger SoA static model, while significantly reducing the communicated data rate by 95\%. Further, our depth models, used to extract intermediate features needed for the downstream task of navigation, performed with up to 81 mean absolute percent error when constructing 2D depth maps from a monocular RGB camera. These models were split and outperformed JPEG compression, which would otherwise communicate the entire image directly to the edge server rather than computing part of the DNN onboard the drone. Thus we have a presented a dynamic multi-branch split DNN for efficient distributed autonomous navigation, along with accurate depth map estimation from a monocular camera.

\section*{Acknowledgment}
This work was supported by National Science Foundation grants CCF 2140154, CNS 2134567, and DUE 1930546.

\bibliographystyle{IEEEtran}
\bibliography{bibliography}

\end{document}